\documentclass[preprint,12pt]{elsarticle}
\usepackage{amssymb}
\usepackage{multirow}%
\usepackage{float}
\usepackage{amsmath,amssymb,amsfonts}%
\DeclareMathOperator*{\argmax}{arg\,max}
\usepackage{amsthm}%
\usepackage{mathrsfs}%
\usepackage{amsmath}
\usepackage{manyfoot}%
\usepackage{multicol}
\usepackage{booktabs}%
\usepackage{algorithm}%
\usepackage{algorithmicx}%
\usepackage{algpseudocode}%
\usepackage{listings}%
\usepackage{hyperref}

\journal{Journal of Visual Communication and Image Representation. }

\begin{document}

\begin{frontmatter}

\title{Beyond Images: Adaptive Fusion of Visual and Textual Data for Food Classification}

\author[label1]{Prateek Mittal}
\ead{prateekmittal154@gmail.com}
\author[label2]{Puneet Goyal}
\ead{puneet@iitrpr.ac.in}
\author[label1]{Joohi Chauhan\corref{cor1}}
\ead{joohi@mnnit.ac.in}

\affiliation[label1]{organization={Motilal Nehru National Institute of Technology Allahabad, India}}

\affiliation[label2]{organization={Indian Institute of Technology Ropar, India}}

\cortext[cor1]{Corresponding Author}

\begin{abstract}
This work introduces a novel multimodal food recognition framework that effectively combines visual and textual modalities to enhance classification accuracy and robustness. The proposed approach employs a dynamic multimodal fusion strategy that adaptively integrates features from unimodal visual inputs and complementary textual metadata. This fusion mechanism is designed to maximize the use of informative content, while mitigating the adverse impact of missing or inconsistent modality data. The framework was rigorously evaluated on the UPMC Food-101 dataset and achieved unimodal classification accuracies of 73.60\% for images and 88.84\% for text. When both modalities were fused, the model achieved an accuracy of 97.84\%, outperforming several state-of-the-art methods. Extensive experimental analysis demonstrated the robustness, adaptability, and computational efficiency of the proposed settings, highlighting its practical applicability to real-world multimodal food-recognition scenarios.
\end{abstract}


\end{frontmatter}

\section{Introduction}\label{sec1}

Multimodal classification is a complex, rapidly evolving, and challenging task, with wide applications in fields such as image classification, speech recognition, and computer vision. It involves the integration and analysis of multiple types of data or modalities \cite{cai2019multi, alonso2021multimodal} to make classification decisions. While multimodal approaches offer accuracy and robustness, the presence of multiple modalities increases the computational complexity. The development of robust systems often requires large amounts of data. Data from multiple modalities are often sourced in a manner that leads to random noise in the data. This noise leads to a decrease in the learning capability of the model. 

Food classification has gained significant attention in recent years, owing to the increasing demand for more personalized and healthy food choices. Food classification systems use various sources of information, such as images, text, and nutritional data, to classify foods into different categories, including fruits, vegetables, meats, and dairy products. This information can be used to create personalized meal recommendations, track food intake, and help people manage their diet and health. The integration of multiple types of data in food classification can improve the accuracy and reliability of classification results. By combining these modalities, a more comprehensive understanding of food can be obtained, leading to more accurate and reliable classification results.

To address these challenges, we present a comprehensive study of multimodal food classification using the UPMC Food-101 dataset \cite{wang2015recipe}, one of the largest and most challenging benchmarks in the field. Our approach combines state-of-the-art deep-learning techniques for image and text classification with an innovative fusion method. Specifically, we employ a modified EfficientNet, called Mod-EfficientNet architecture, with Mish activation for image classification, a BERT-based model for text classification, and a dynamic multimodal fusion strategy that adaptively integrates visual and textual features to ensure optimal information utilization while reducing the adverse effects of incomplete or inconsistent multimodal inputs.

The main contributions of our study are as follows: 

\begin{enumerate}
    \item This work introduces a novel dynamic fusion strategy that adaptively integrates visual and textual modalities for food categorization. Unlike traditional early or late fusion approaches, this method ensures optimal feature utilization while mitigating the impact of missing or noisy modalities, thereby achieving state-of-the-art results.
    \item The dynamic fusion technique introduces an adaptive weighting mechanism where the reliance on textual or visual inputs varies based on their certainty levels. This reduces classification errors caused by missing or ambiguous data, making it a more applicable real-world solution for food recognition.
    \item The paper proposes a modified EfficientNet with Mish activation for image classification and BERT-based deep learning models for text classification. The combination of these architectures significantly outperforms unimodal approaches, improving classification robustness despite the high data noise in the UPMC Food-101 dataset.
    \item The proposed fusion model achieves 97.84\% accuracy, outperforming prior best-performing models by over 4\%. It surpasses transformer-based multimodal classifiers and CNN-text hybrid architectures, thereby setting a new benchmark for multimodal food categorization research.
    \item The proposed method is rigorously evaluated on UPMC Food-101 and UPMC Food-25, which contain highly noisy, incomplete, and unstructured data. This study systematically investigated the effect of data quality on the classification performance, highlighting the advantages of multimodal learning.
    
\end{enumerate}

The remainder of this study builds upon the concepts and ideas introduced in this section. In Section \ref{sec:Related_Work}, the contribution of the existing literature is discussed in detail, along with the research gaps. Section \ref{sec:Methodology} describes the architecture of the model, which was finally selected as the best-performing method. It also contains subsections on Image Classification, Text Classification and Fusion techniques. Section \ref{sec:Experimental_Setup_and_Results} consists of six subsections: Dataset Description, Unimodal Experiments (further divided into Classification of Textual Data and Classification of Image Data), Data Stratification, Domain Adaptation on Stratified Data and Fusion of Image and Text Modalities. In Section \ref{sec:Result_Analysis_and_Discussions} we discuss the experimental results in detail, along with possible explanations of some observations. Finally, Section \ref{sec:Conclusion} concludes the paper. 

\section{Related Work}
\label{sec:Related_Work}
Multimodal food classification, a subfield of computer vision, aims to leverage the complementary nature of image and text data to improve food recognition accuracy. Wang et al. \cite{wang2015recipe} presented the dataset UPMC Food-101, which consists of images and text. They also demonstrated the viability of multimodal food classification using separate models for image (VGG19) and text (TF-IDF) data with late fusion to combine the predictions. This approach established the foundation for future research but lacked robust joint representation learning of image and text features. Gallo et al. \cite{gallo2018image} and Nawaz et al. \cite{nawaz2018learning} proposed incorporating text features directly into the image domain to create information-enriched images. While achieving some improvement, these methods struggle to capture the complex relationships between the visual and textual representations of food. Narayana et al. \cite{narayana2019huse} proposed Hierarchical Universal Sentence Embedding (HUSE), a method that utilizes strong uni-modal architectures (BERT for text and Graph-RISE for images) with late fusion. This approach paved the way for modern techniques that build strong individual models before leveraging their combined strengths for improved classification. Another approach for developing multimodal architectures is to use the importance of each modality as well as the relationship between various modalities to develop an accurate joint representation of the features extracted from each modality. De La Comble et al. \cite{de2022multi} and Im et al. \cite{im2021cross} explored assigning weights to modalities and hierarchical training strategies to account for the importance and relationships of the modality. 

Ak et al. \cite{ak2023leveraging} propose a multimodal classification method that uses transformers to efficiently train and fuse features. Their work focused on leveraging the strengths of the transformers for multimodal tasks. Mengmeng Ma et al. \cite{Ma_2022_CVPR} presented a strategy to improve the robustness of transformer models. They highlighted the limitations of transformers, which are powerful deep learning architectures, in handling missing modalities. Their work emphasized the need for robust fusion strategies, even with powerful base models. Unlike traditional methods that focus on aggregating information from different modalities, UniS-MMC \cite{zou2023unis} leverages a contrastive learning approach to obtain more reliable multimodal representations. It exploits weak supervision from unimodal predictions and trains the model to improve agreement among these unimodal representations. Wajid et al. \cite{wajid2024deep} presented a deep learning approach for multimodal classification that incorporates neutrosophy. It addresses this task by transforming text information into an image format and then combining it with the original image data. A neutrosophic convolutional neural network (NCNN) is then employed to learn feature representations from this combined data for classification purposes.

Kiela et al. \cite{kiela2018efficient} explore multimodal fusion methods for large-scale classification tasks, achieving high accuracy with techniques like additive fusion and max pooling. Their work highlighted methods for efficient fusion, including the discretization of visual features. Li et al. \cite{li2023efficient} focused on efficient information integration and explored interactive prompting techniques to improve the fusion between text and image data for tasks such as classification. Mao et al. \cite{mao2021visual} presented a two-step framework for food categorization that includes food localization and hierarchical food classification. These methods use convolutional neural networks (CNNs) as the backbone architecture, and clustering is performed to generate a hierarchical structure that encapsulates the semantic visual relations among food categories. Recently, Saklani et al. \cite{saklani2024ameliorating} used traditional state-of-the-art text and image classifiers on a modified UPMC Food-101 dataset with preprocessing. This study mainly focuses on benchmarking with data preprocessing.

There have also been some developments around the use of stacked learning approaches for the classification of the UPMC Food-101. Suresh and Verma \cite{suresh2024stacking} proposed two major voting ensemble models that outperformed the state-of-the-art level-0 CNN models: EfficientNetV2S, InceptionV3, and Fusion – LSTM + InceptionV3. They also proposed seven metalearner stacking models based on three level-0 models. The best performing stacking model was the SVM linear stacking model, which achieved 93.1\% classification accuracy. 

Pan et al. \cite{pan2024fmifood} have explored vision transformers \cite{dosovitskiy2020image} as the unimodal baseline and various methods for multimodal contrastive learning to obtain a final classification result on the UPMC Food-101 dataset. CLIP \cite{radford2021learning} has been used for contrastive learning on image-text pairs to make matched pairs as close as possible while separating unmatched pairs based on global features. FILIP \cite{yao2021filip} is another model that the authors have experimented with which works similarly to CLIP but conducts contrastive learning based  on fine-grained relationships using image patch features and text token features. The authors also experimented with UniCL \cite{yang2022unified}, a model that considers multiple texts matched to an image within a batch during a contrastive learning process. The authors also experimented with iCLIP \cite{wei2023iclip}, which is a CLIP model applied to image classification tasks with augmented text descriptions from the WordNet dictionary. The text input consists of the text label along with its corresponding explanation from the WordNet dictionary. They utilized GPT-4 for text augmentation. The authors have also proposed a new model, FMiFood \cite{pan2024fmifood}, a specially designed model for multimodal food categorization, which  introduces a flexible matching mechanism that allows an image patch to match multiple text tokens or none, as appropriate.

Current literature explores various methods for classifying food images using single and multiple modalities. However, it can be consistently observed that the multimodal approach always outperforms the unimodal approaches. We can also conclude from the literature that although attempts to improve the results have been made by various researchers, there are limited or no articles where strong analysis has been conducted. Taking inspiration from these facts, we explored various image classification, text classification, and fusion techniques along with robust analysis to promote high standards of transparency and reproducibility of results, which would enable end users to utilize the results of the research more effectively.

\section{Methodology}
\label{sec:Methodology}

Multimodal food classification can be formalized as a supervised learning problem in which the goal is to learn a mapping function f(x) that takes a set of features x as input and maps them to a set of food categories y. Features x can be multimodal, such as images, text, or audio features, and food categories y can be represented as a one-hot encoded vector or probability distribution over a set of food categories.

Mathematically, this problem can be represented as follows.

Given \({(x_1, y_1), (x_2, y_2), ..., (x_N, y_N)} \) a set of N training samples, where \(x_i\) is a set of features for a food item, and \(y_i\) is the corresponding food category, we find a function \(f(x)\) that minimizes the following loss function L:
\begin{equation}\label{main_loss}
L = \frac{1}{N} \sum_{i=1}^{N}(y_i - f(x_i))^2 
\end{equation}

where \(y_i\) is the true label and \(f(x_i)\) the predicted label for the \(i^{th}\) food item. The loss function measures the difference between the true label and the predicted label, and the goal is to minimize the loss function by determining the optimal function f(x).

In multimodal food classification, the function \(f(x)\) takes as input a set of multimodal features \(x\) and maps them to a set of food categories \(y\), allowing the model to make predictions based on multiple sources of information. The overall architecture of the proposed method is shown in Figure 1.

\begin{figure}
\centering
\includegraphics[scale=0.50, keepaspectratio]{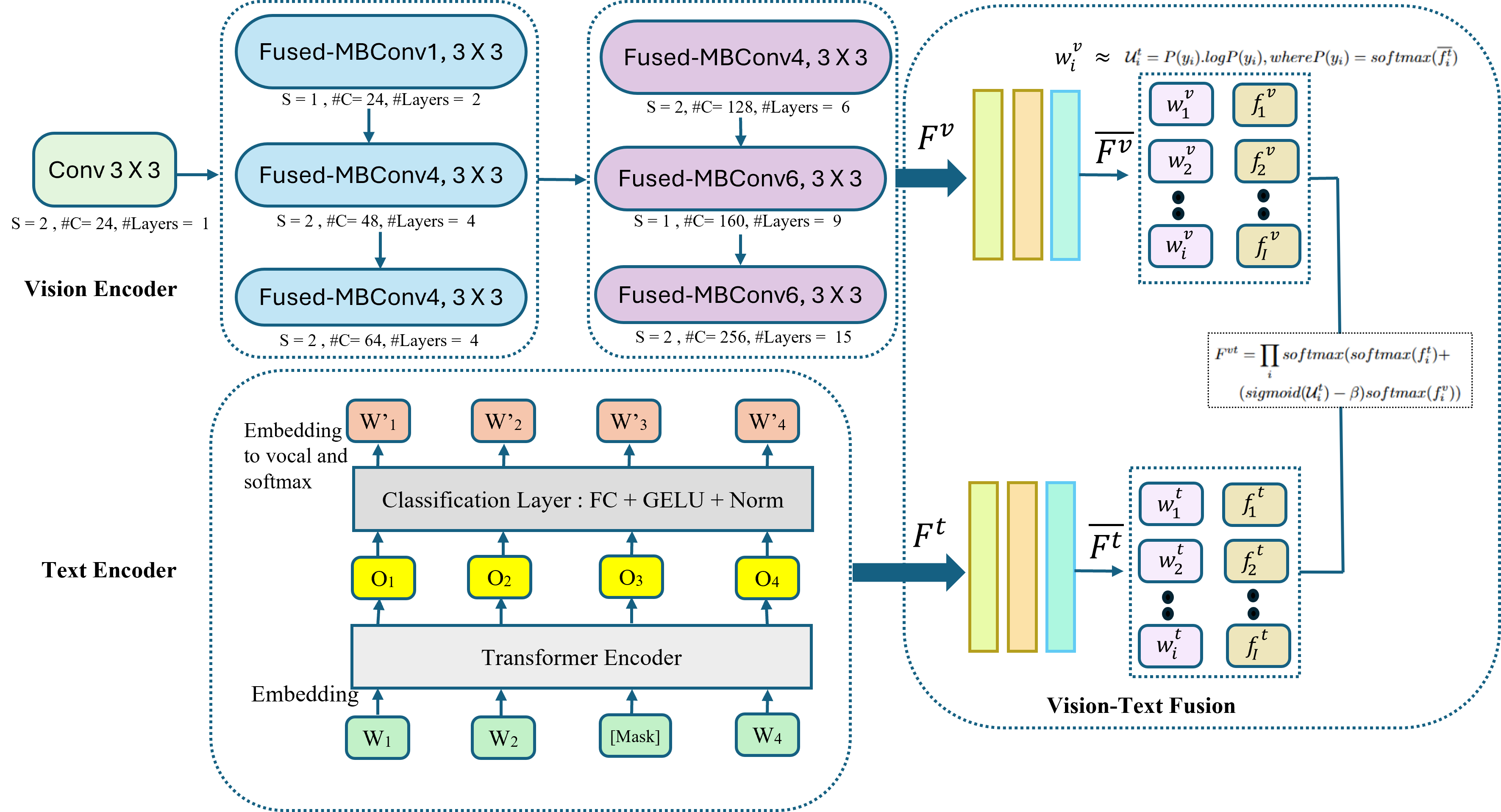}
\caption{Overview of the proposed network. The network comprises; an image encoder for extracting the visual features; a text encoder for extracting textual features; and a dynamic fusion module that extracts the complementary information from multiple modalities and integrates the feature maps for the final prediction task.}
\label{architeture-efficient}
\end{figure}

\subsection{Image Classification}

The proposed approach consists of a unimodal CNN network that corresponds to images. This image classifier is one of the most powerful convolutional networks, EfficientNet \cite{tan2019efficientnet}. However, it has been reported that the training of EfficientNet on a large dataset is quite slow because of the memory usage of large images \cite{tan2019efficientnet}. Therefore, to speed up the training process and make the model parameter-efficient, FixRes\cite{tan2021efficientnetv2} was added to EfficientNet. In general, there are eight stages in the CNN network, and instead of using only Mobile inverted bottleneck convolution (MBConv) \cite{sandler2018mobilenetv2}, the network has the combination of Mobile inverted bottleneck convolution (MBConv) and fused-MBConv \cite{tan2021efficientnetv2} in the initial stages of the model. The output feature map from the image classifier is denoted as \(F^v\).

Fused-MBConv increases the number of parameters and FLOPs; therefore, it is not desirable to use it in all stages of the network. In general, depthwise convolutions are much slower in the initial stages and speed up in the later stages; therefore, to improve the training speed, fused-MBConv is used in the initial stages of the network and MBConv in later stages. Furthermore, a smaller expansion ratio is used in fused-MBConv and MBConv, that is, \{1,4\} and \{4,6\}, respectively, to reduce the memory overhead. In Mod-EfficientNet, to extend the network capacity without introducing runtime overhead, a non-uniform scaling strategy is introduced in the proposed network (Equation \ref{eq:1}) that gradually scales up and adds more layers to the later stages of the network.

\begin{equation} \label{eq:1}
c = \alpha^\delta ;\\
d = \beta^\phi  ;\\
w = \gamma^\zeta ;\\ 
\end{equation}
where \(\alpha^2 . \beta . \gamma^2 \approx 2\) and
\(\alpha, \beta, \gamma \geq 1\), \(\delta, \phi, \zeta\) are the compound coefficient corresponding to each dimension, \(\alpha, \beta, and \  \gamma\) are the constants responsible for managing the distribution of the available resources identified by the compound coefficient to all the network dimensions.

We used a self-regularized nonmonotonic activation function, that is, Mish \cite{misra2019Mish}. Mish has been shown to integrate well with modern neural network architectures, including those that employ normalization layers like Batch Normalization. The smooth nature of the Mish complements the normalizing effects, leading to more stable training dynamics. This function lies in the range of $[\approx-0.31, \infty]$. In addition, it utilizes the self-gating property, in which the output of a nonlinear function is multiplied by a non-modulated input. Mathematically, the Mish activation function is \(\Psi(x)\), where \(x\) is a variable that represents the input value.

\begin{equation} \label{mish:definition}
\Psi(x) = xtanh(softplus(x)) = xtanh(ln(1+e^x))
\end{equation}

Mish is a smooth, non-monotonic activation function with continuous first and second derivatives (Equation \ref{mish:derivative} and \ref{mish:second_derivative} respectively). The smoothness of Mish can be crucial for the stability and performance of deep learning models because it avoids discontinuities or sharp transitions that can cause optimization issues, particularly in gradient descent.

\begin{equation} \label{mish:derivative}
   \frac{d\Psi(x)}{dx} = \frac{e^{x} \omega}{\delta^{2}} \\
\end{equation}

Where,

\[  \omega = 4(x+1) + 4e^{2x} + e^{3x} + e^{x}(4x+6) \]
\[ \delta = 3e^{2x} + 2 \]

\begin{equation} \label{mish:second_derivative}
    \frac{d^{2}\Psi(x)}{dx^{2}} = \frac{2\mathrm{e}^{x} \Lambda}{\delta^{2}}
\end{equation}

Where, 

\[ \Lambda = 3{e}^{5x} + 6\mathrm{e}^{4x} + 12e^{2x} + 10\mathrm{e}^{x} - 6x + 8 \left(x + 2\right) \mathrm{e}^{x} + 4x + 8 \]

Mish adapts to the input data dynamically because of its nonlinearity and smooth gradient, potentially leading to better generalization and robustness in various learning tasks, especially in deep neural networks where gradient flow is critical.

Therefore, our Image Classification model first uses a modified EfficientNet model combined with the Mish activation function to extract a feature map from the images. 

\begin{equation}
    \chi = Mod-EfficientNet(I^v) \\
\end{equation}

\begin{equation}
     F^v = \Psi(\chi) \\ 
\end{equation}

where \(I^{v}\) denotes the input set of images, \(\chi\) denotes the initial feature map returned by Mod-EfficientNet, \(\Psi\) denotes the Mish activation function as defined in Equation \ref{mish:definition} and \(F^{v}\) denotes the final image feature map. 

\subsection{Text Classification}
The task of text classification is handled by Bidirectional Encoder Representation of Transformers (BERT) \cite{devlin2018bert}. The key innovation of BERT is the application of bidirectional transformer training to language modeling.  BERT uses transformer \cite{vaswani2017attention}, which is an attention mechanism that learns contextual relations between words (or sub-words) in a text. This transformer is composed of two components: an encoder and decoder. The encoder is responsible for reading the input text and the decoder is responsible for generating predictions. The Transformer encoder is designed such that it reads the entire sequence of words simultaneously. Therefore, it is considered to be bidirectional. This trait of the model allows it to interpret the context of a word in the domain of all neighboring words.
In the encoder phase, a sequence of tokens (which are the inputs) is first embedded into the vectors and then processed in the neural network. A sequence of vectors of size \(H\) is obtained as the output. Each vector in the output sequence corresponds to an input token with the same index. Before providing the word sequences as input to BERT, 15\% of the words were replaced with a [MASK] token. Based on the context provided by the non-masked words in the sequence, the model attempted to predict the original value of the masked words. Finally, we calculated the probability of each word in the vocabulary using Softmax. The feature map obtained from the text encoder is represented as \(F^t\) (Equation \ref{eq:text_extraction}) 

\begin{equation} \label{eq:text_extraction}
    F^t = BERT(I^t)
\end{equation}

where \(I^t\) denotes a set of text inputs.

\subsection{MultiModal fusion}\label{subsec:Multimodal_Fusion}
The multimodal fusion module aims to integrate the characteristics of textual and visual features, \(F^t\), \(F^v\), to efficiently extract complementary features across multiple modalities. The main challenge in multimodal fusion is to capture cross-modal correlations accurately and enable flexible cross-modal interactions. Typically, there are two approaches for learning joint representations from multiple modalities, that is, late fusion and early fusion, as illustrated in Figure 2. To fully leverage the potential of each modality, while minimizing the impact of low-quality multimodal data, we adopted a dynamic multimodal fusion approach.

\begin{figure}[!ht]
\centering
\includegraphics[scale=.5, keepaspectratio]{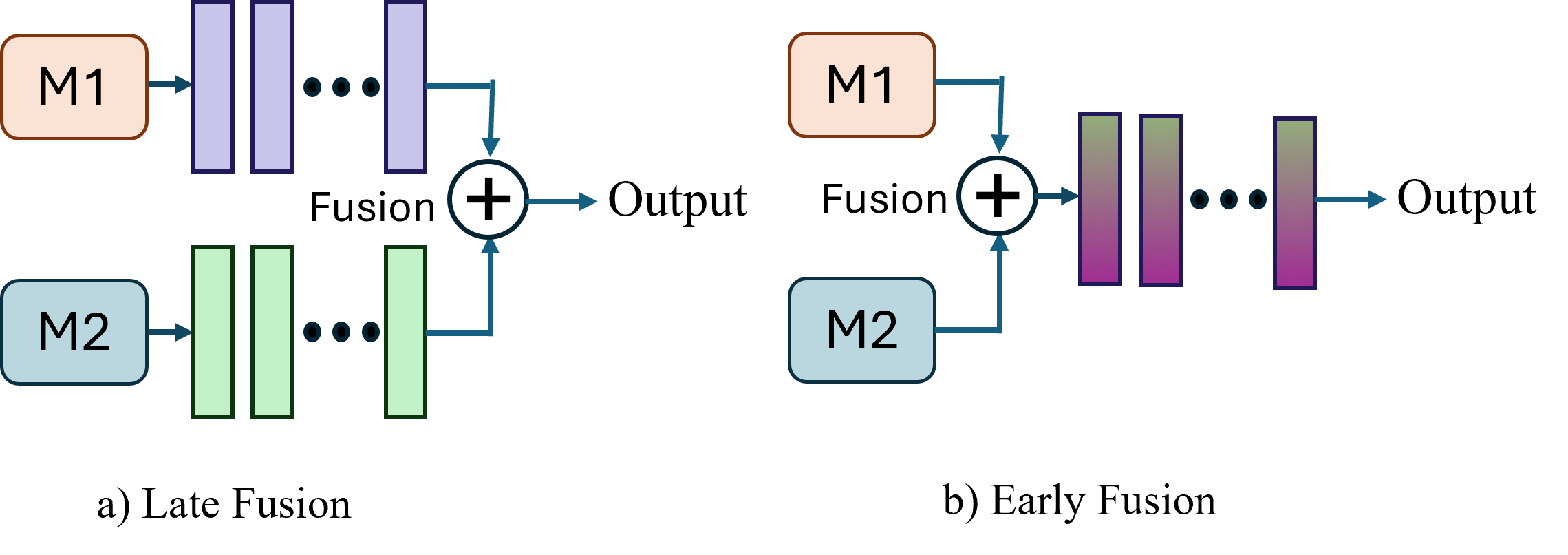}
\caption{Illustrations of commonly used multimodal networks. These approaches learn a joint representation with fusion mechanisms (late or early) from the embeddings of modality 1 and 2.}	
\end{figure}

Initially, to remove the feature map size restrictions, feature up-sampling is performed to uniformly up-sample the image and text feature vector by adding a \(256 \times 256\) dense layer and a ReLu activation function. Further, each output feature map is passed through the Global Average pooling 2D (GAP) layer and modality-wise fusion is performed later. 
\begin{equation}
\begin{split}
\overline{F^t} & = ReLu(W^tF^t + b^t) \in R^d \\
\overline{F^t} & = GAP(\overline{F^t})  
\end{split}
\end{equation}
\begin{equation}
\begin{split}
\overline{F^v} & = ReLu(W^vF^v + b^v) \in R^d \\
\overline{F^v} & = GAP(\overline{F^v})
\end{split}
\end{equation}
where \(W^t\) and \(W^v\) represent the weights, and \(b^t\) and \(b^v\) are the respective biases. Here, \(R^d\) denotes the set of features, each with a \(d\) dimension representing the number of units in the dense layer.

The fusion of modalities is based on element-wise addition of the respective feature maps, and a dynamic weight factor is added to each \(i^{th}\) step:

\begin{equation}
    f^{vt}_i = w^t_i.{\overline{f^t_i(X)}} + w^v_i.{\overline{f^v_i(X)}} 
\end{equation}

Theoretically, the generalization error of the multimodal classifier is bound by the weighted average performance of all unimodal classifiers in terms of empirical loss, model complexity, and the covariance between the fusion weight and unimodal loss. A static fusion strategy leads to a higher upper bound of the generalization error compared with dynamic fusion. In this study, we focus on fusing visual information to address data uncertainty in the learning process of the text encoder. The uncertainty of a \(y_i\) by the text encoder is given by the entropy of the predictive posterior

\begin{equation}
{\mathcal{U}^t_i} = P(y_i).logP(y_i), where P(y_i) = softmax(\overline{f^t_i})
\end{equation}
A large \({\mathcal{U}^t_i}\) indicates large uncertainty when the text encoder predicts the current token, which requires more visual compensation from \({\overline{f^v_i}}\). Because we fuse the two modalities and consider the predominant role of the text encoder within them, we set \({w^t_i}\) to 1 and dynamically modulate \({w^v_i}\) in terms of \({\mathcal{U}^t_i}\). Therefore, the uncertainty-based dynamic fusion can be represented as

\begin{equation}
F^{vt} = \prod_i softmax(softmax({f^t_i}) + (sigmoid({\mathcal{U}^t_i}) -\beta) softmax({f^v_i}))
\end{equation}

where \(\beta\) denotes a hyperparameter with a default value of 0.5. Note that if the text encoder is extremely confident, that is, \(({\mathcal{U}^t_i} \rightarrow 0^+)\), then the final decision can be completely based on the decision of the text encoder \((sigmoid({\mathcal{U}^t_i}) - 0.5 \rightarrow 0^+)\). Otherwise, the weight of the vision encoder increased with an increase in \({\mathcal{U}^t_i}\). This fusion module performs a fusion operation between the visual and text features, and the final output feature map can be fed into a classifier for the final prediction task. Figure \ref{architeture-efficient} depicts the entire process of extraction of image and text features, as well as the adaptive fusion mechanism, which helps in learning the complementary features from all modalities. 

\subsection{Final Classification}
After the fusion of text and image feature maps using the strategy described in Section \ref{subsec:Multimodal_Fusion}, we used the softmax classifier to obtain normalized classwise confidence scores. Let the correct class distribution be \(p\) (where the probability of the correct class is 1 and the rest are 0) and the estimated distribution given by the softmax function be \(q\). Then, if there are \(c\) number of classes the cross-entropy loss between \(p\) and \(q\) is given by Equation \ref{eq:cross_entropy_loss}

\begin{equation}\label{eq:cross_entropy_loss}
    \upsilon(p,q) = - \sum_{i=1}^{c} p_{i} log(q_{i})
\end{equation}

In our case, under the assumption that \(F^{vt}\) has the same dimension as the number of classes (after flattening),  the distribution \(q\) is given by Equation \ref{eq:distribution_q} ( The Softmax function)

\begin{equation} \label{eq:distribution_q}
    q_{i} = \frac{e^{F^{vt}_{y_i}}}{\sum_{j}e^{F^{vt}_{j}}}
\end{equation}

where \(F^{vt}_{j}\) denotes the \(j^{th}\) element of the class scores \(F^{vt}\) and all other symbols have the meaning defined earlier in this section.

The softmax function takes a vector of real-valued scores as input and transforms it into a vector of values between 0 and 1, which sum to 1. Softmax classifier is minimising the cross entropy between the estimated distribution \(q\) (obtained by softmax function) and the actual distribution \(p\). Because \ref{eq:cross_entropy_loss} can also be written in terms of entropy and the Kullback-Leibler divergence (Equation \ref{cross_entropy_modified}), and the entropy of \(p\) is 0 ( which means that \(H(p) = 0 \)), minimizing the cross-entropy loss function is equivalent to minimizing the Kullback-Leibler divergence between the two distributions, which means it minimizes the difference between the actual distribution \(p\) and the estimated distribution \(q\). Hence, the cross-entropy objective requires the predicted distribution to have all of its masses in the correct class. 

\begin{equation}\label{cross_entropy_modified}
    \upsilon(p,q) = H(p) + D_{KL}(p\parallel q)
\end{equation}

While practically implementing the softmax function, the intermediate terms \(e^{F^{vt}_{y_i}}\) and \(e^{F^{vt}_{j}}\) may become very large, causing unexpected errors owing to the division of large numbers. Therefore, to control for these unexpected errors, we introduce a regularization term \(\eta\) (Equation \ref{softmax_modified} ).

\begin{equation}\label{softmax_modified}
    \frac{e^{F^{vt}_{y_i}}}{\sum_{j}e^{F^{vt}_{j}}} = \frac{\eta e^{F^{vt}_{y_i}}}{\eta \sum_{j}e^{F^{vt}_{j}}} = \frac{e^{F^{vt}_{y_i} + log(\eta)}}{\sum_{j}e^{F^{vt}_{j} + log(\eta)}}
\end{equation}

where all symbols have their usual meaning. 

The regularization term \(\eta\) can be selected randomly and, as evident from Equation \ref{softmax_modified}, it does not cause any change in the numerical output of the softmax function but improves the numerical stability of the computation. We require the highest value inside vector \(F^{vt}\) to be zero. Therefore, we choose \(log(\eta) = -max_{j}F^{vt}_{j}\). This simply means that all values inside the vector \(F^{vt}\) should be shifted such that the highest value is 0. 

The final distribution for obtaining the final output is generated by minimizing the cross-entropy loss, as defined in Equation \ref{eq:cross_entropy_loss}. The minimization process is performed using the gradient update rule and continues until a fixed number of iterations (whichever occurs earlier) or until \(\upsilon(p,q) \leq \epsilon\) where \(\epsilon\) is an arbitrarily small threshold. We assume that the final distribution of class probabilities obtained using this method is given by \(\theta\). The final output of the classifier is then given by Equation \ref{eq:final_classification}.

\begin{equation} \label{eq:final_classification}
    f(x_{i}) =  \argmax \theta
\end{equation}

where \(x_i\) is the set of input features (combination of image and text input in our case) and \(f(x_{i})\) is the function defined in Equation \ref{main_loss}. It is used to represent the output class as calculated by the final classification layer. These outputs are then compared with the actual class labels \(y_{i}\), and the training is continued until the loss converges or satisfactory performance is obtained. 

\section{Experimental Setup and Results}
\label{sec:Experimental_Setup_and_Results}
\subsection{Dataset Description}
The proposed network and the other state-of-the-art methods are evaluated on a large open-source dataset “UPMC Food-101” \cite{wang2015recipe}. This dataset is one of the benchmarks for multimodal architectures and frameworks, and each data sample is a combination of image-text pairs.

\begin{figure*}
	\centering
		\includegraphics[scale=.45, keepaspectratio]{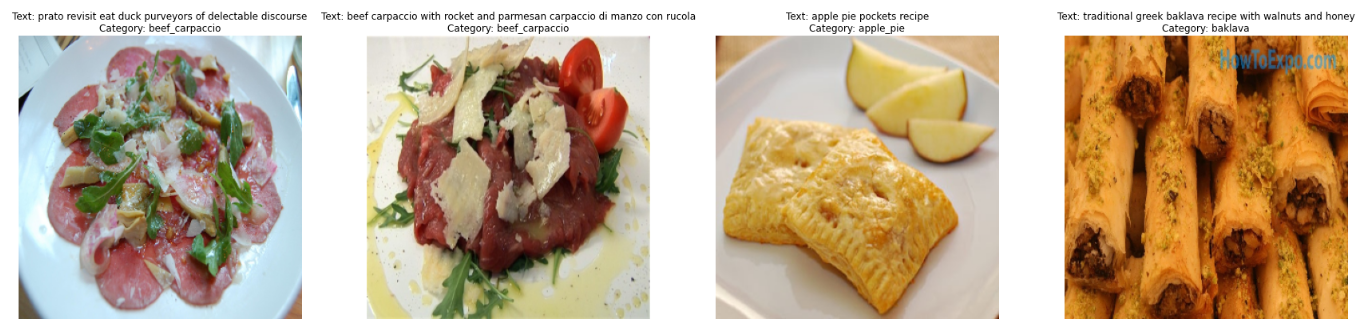}
	\caption{Sample images from the UPMC Food- 101 dataset along with their associated textual description and categories.}
	\label{data-samples}
\end{figure*}

The data for UPMC Food-101 were gathered in an uncontrolled manner, resulting in a very noisy dataset, thus making it more challenging. It consists of 90,704 images with their related text, out of which 67,988 images were used for training and 22,716 were used in the test dataset. Each image was classified into one of 101 food categories. This dataset shares its characteristics with those of ETHZ 101 [10]. The difference between the two datasets lies in the compilation and structure methods. Figure \ref{data-samples} shows some of the images from the UPMC Food-101 dataset, along with the related texts and labels.

For the initial validation of the proposed approach, considering the challenges and the large dataset size, we shortened the original dataset of 101 classes to 25 classes. In the shortened dataset “UPMC Food-25, ” the first 25 classes were selected from 101 classes. The original dataset consisted of 90,704 images distributed in 101 classes, whereas the shortened dataset consisted of 22,502 images distributed into 25 classes. The training set of UPMC Food-25 consisted of 16,867 images, and the remaining 5635 images were used to test the model performance. 

\subsection {Unimodal Classification Experiments}

The deep learning models for image classification were pre-trained on the ImageNet dataset as the backbone of the CNN, and the transformer networks for textual classification were pre-trained on BooksCorpus and English Wikipedia. All experiments were conducted using four parallelly computed 16GB NVIDIA Quadro RTX5000 GPUs. The codes were written using TensorFlow and Keras. The hyperparameters used for all the experiments are described and listed wherever necessary. The learning rate was fixed during the training process, and the hyperparameters were computed using the validation set. The models were trained for 50 epochs until loss converged. 

\subsubsection{Classification of Textual Data}
For text classification, we tried both machine learning and deep learning-based frameworks. In the machine learning approach, we utilized the Scikit Learn Library to implement the selected algorithms, and a count vectorizer and TF-IDF Transformer were used on the original textual data. The results of these transformations are then fed into traditional machine learning networks, such as multi-nominal naive Bayes, stochastic gradient descent (SGD), and Logistic Regression (LR). Based on the preliminary results obtained through these networks, we further implemented a voting-based ensemble of multinominal naive Bayes, SGD, and LR with a voting ratio of 2:2:1 in the respective order. Moreover, for efficacy, deep learning and transformer models were implemented and imported from the hugging-face library. For our experiments, the batch size was $64$ for all the selected transformer-based models. We used Sparse Categorical Cross-Entropy as our objective for the cost function and Adam optimizer function. The learning rate is set to \(7.5 \times e^{-5}\) throughout the training process.

The results for the textual part of the UPMC Food-25 and UPMC Food-101 datasets are presented in Tables \ref{Shortened-text}  and \ref{original-text}, respectively. From the results, it can be observed that for both UPMC Food-101 and UPMC Food-25, the Voting Ensemble method performed better in machine learning-based approaches, and BERT outperformed the other deep learning-based approaches for text classification.

\begin{table}
\caption{Performance obtained by machine learning and deep learning techniques on the UPMC Food-25 dataset for text classification.}
\vspace{0.1in}
\label{Shortened-text}
\resizebox{\textwidth}{!}{
\begin{tabular*}{445pt}{cccccc}
\toprule
& Model Name & Accuracy & F1 & Precision & Recall\\
\midrule
Machine Learning & SGD & 0.847 & 0.874  & \textbf{0.926} & 0.847 \\
                 & Multinomial Naive Bayes & 0.855 & 0.854 & 0.857  & 0.855 \\
                 & Logistic Regression & 0.869 & 0.870 & 0.873  & 0.869  \\
                 & Voting Ensemble  & \textbf{0.875} & \textbf{0.877} & 0.882  & \textbf{0.875}  \\
\midrule
Deep Learning & XLNet & 0.868 & 0.869 & 0.869 & 0.879  \\
                 & RoBERTa & 0.869 & 0.872 & \textbf{0.873} & 0.879 \\
                 & BERT & \textbf{0.874} & \textbf{0.875} & 0.870 & \textbf{0.880} \\

\bottomrule
\end{tabular*}
}
\end{table}

\begin{table}
\centering
\caption{Performance obtained by machine learning and deep learning techniques on the UPMC Food-101 dataset for text classification.}
\vspace{0.1in}
\label{original-text}
\resizebox{\textwidth}{!}{
\begin{tabular*}{445pt}{cccccc}
\toprule
& Model Name & Accuracy & F1 & Precision & Recall\\
\midrule
Machine Learning & SGD & 0.826 & 0.860 & \textbf{0.916} & 0.826\\
                 & Multinomial Naive Bayes & 0.833 & 0.833 & 0.84 & 0.833 \\
                 & Logistic Regression & 0.852 & 0.853 & 0.855 & 0.852 \\
                 & Voting Ensemble  & \textbf{0.859} & \textbf{0.859} & 0.862 & \textbf{0.858} \\
\midrule
Deep Learning  & RoBERTa & 0.849 & 0.848 & 0.854 & 0.853 \\
                 & XLNet & 0.858 & 0.859 & 0.857 & 0.863  \\
                 & BERT & \textbf{0.884} & \textbf{0.865} & \textbf{0.863} & \textbf{0.872} \\

\bottomrule
\end{tabular*}
}
\end{table}

\subsubsection{Classification of Image Data}

Initially, data augmentation techniques were used to augment the dataset, and the augmented images were fed as inputs to a pre-trained model. Experiments were performed with different learning rates fixed throughout the training, that is, \(1e^{-5}\) and \(7.5e^{-5}\), whereas the batch size was fixed to 32 because of constraints on computational resources. 

The results obtained using different image classification models with different network settings on the UPMC Food-101 and UPMC Food-25 datasets are presented in Tables \ref{original-image} and \ref{shortened-image}, respectively.  From Table \ref{original-image} and Table \ref{shortened-image}, it can be noted that the modified(Mod)-EfficientNet, when used with the Mish activation function, performs better with a learning rate of \(7.5e^{-5}\) for both the UPMC Food-101 and UPMC Food-25 datasets.

\begin{table}
\centering
\caption{Performance (Accuracy) by the proposed approach and the selected image classification methods on UPMC Food-101 dataset}
\vspace{0.15in}
\label{original-image}
\begin{tabular}{ccc}
\toprule
Model Name &  Learning Rate = \(7.5e^{-5}\) & Learning Rate = \(1e^{-5}\) \\
\midrule
VGG-16 & 0.258 & 0.242\\
ResNet152V2 & 0.522 & 0.500\\
ResNet101 &  0.597 & 0.588\\
InceptionResnetV2 & 0.638 & 0.602 \\
EfficientNet B7 & 0.651 & 0.621 \\
Mod-EfficientNet & 0.683 & 0.636\\ 
Mod-EfficientNet + 'Mish' & \textbf{0.736} & \textbf{0.697} \\
\bottomrule
\end{tabular}
\end{table}

\begin{table}[]
    \centering
    \caption{Classification accuracy of Mod-EfficientNet with ReLU and Mish activation functions under different learning rates (LR) on the UPMC Food-25 dataset. Best results for each LR are highlighted in bold.}
    \vspace{0.1in}
    \label{shortened-image}
    \begin{tabular}{cccc}
    \toprule
    Model Name & Activation Function & LR=\(7.5e^{-5}\) & LR= \(1e^{-5}\) \\
    \midrule
    Mod-EfficientNet & ReLU & 0.750 & 0.722 \\
    Mod-EfficientNet & Mish &  \textbf{0.789} & \textbf{0.741} \\
    \bottomrule
    \end{tabular}
\end{table}

\subsection{Data Stratification}

 We aimed to capture the most accurate and detailed features of the dataset. In the experiments described in the previous sections, we observed that the randomness of various classes resulted in lower precision and lower accuracy. This led us to believe that some classes might have more noise than others, and hence, the overall accuracy was impacted. To mitigate this problem, we divided the entire dataset of 101 classes into three groups: poor, average, and best. The representativeness of each class was used as the criterion for dividing the data into classes. Based on the precision of classification using the EfficientNet-Mish model, the 101 classes were divided into poor, average, and best classes (as mentioned above). The distributions of the classes are listed in Table \ref{division}.

\begin{table*}
\centering
\caption{Grouping of food classes in the UPMC Food-101 dataset based on classification performance. Groups are defined by Mean F1-score with corresponding standard deviations across classes in each group.}
\vspace{0.1in}
\label{division}
\begin{tabular}{cccc}
\toprule
Group Name &  Number of Classes & Mean F1 & Standard Deviation \\
\midrule
Poor Classes    & 15  & 0.4573 & 0.0683\\
Average Classes & 54 & 0.6403 & 0.0504 \\
Best Classes    & 32 & 0.7262 & 0.0546 \\
\bottomrule
\end{tabular}
\end{table*}

We first tested the stratified data using Effmish (the best-performing model for a given dataset). The results of this test are presented in Table \ref{initial_stratified}. From these results, we observed that the reason for the lower accuracy on the complete dataset may be that the combination of all classes hampers the representation of other classes and does not allow the model to learn the features effectively. 

\begin{table*}
\centering
\caption{Performance comparison of the proposed method on stratified data across different class groups. Improvements are reported over previous Mean F1-scores, with Top-5 accuracy included for additional insight.}
\vspace{0.1in}
\label{initial_stratified}
\resizebox{\textwidth}{!}{
\begin{tabular*}{450pt}{ccccc}
\toprule
Group Name & Mean F1 & Mean F1 (Previous) & Improvement & Top-5 Accuracy\\
\midrule
Poor Classes    & 0.6341 & 0.4573 & 0.1768 & 0.8466  \\
Average Classes & 0.6812 & 0.6403 & 0.0409 & 0.8485   \\
Best Classes    & 0.8103 & 0.7262 & 0.0841 & 0.9248   \\
\bottomrule
\end{tabular*}
}
\end{table*}

\subsection{Domain Adaptation on Stratified Data}

In our experiments, we used the weights obtained during the training of the best classes for the domain adaptation task. That is, the weights obtained during the training of the best classes were used to learn the effective and accurate feature representations of the poor and average classes. After domain adaptation, the training of the poor classes plateaued at the seventh epoch. The main experimental benefit of Domain Adaptation over normal-phase training is that the model converges earlier. Additionally, we obtained a mean F1 score of 0.8103. In other words, an improvement of 0.0841 in the mean F1 score was observed over the initial mean F1 score. The Top five accuracies obtained using this method were 0.9248.

\subsection{Fusion of Image and Text Modalities}

Using the above experiments, we selected the best-performing image and text classification models and compared various fusion techniques for combining the feature vectors received from these two modalities. These fusion techniques are compared in Table \ref{fusion-ablation}.  From these results, it can be clearly observed that the proposed dynamic fusion method outperforms the other methods as well as the traditional early and late fusion techniques.

\begin{table}
\centering
\caption{Ablation study of different fusion strategies (Late, Early, Adaptive) applied to image and text modalities on the UPMC Food-101 dataset. Results report individual modality performance and overall fusion accuracy, with best results highlighted in bold.}
\vspace{0.1in}
\label{fusion-ablation}

\begin{tabular}{cccc}
\toprule
Model Name & Image & Text & Fusion \\
\midrule
 Late Fusion & 73.60 & 88.42 &  93.56 \\
 Early Fusion & 73.60 & 88.42 & 94.98 \\
 Adaptive Fusion  & 73.60 &  \textbf{88.42}  & \textbf{97.84}  \\
\bottomrule
\end{tabular}
\end{table}

\section{Result Analysis and Discussion}
\label{sec:Result_Analysis_and_Discussions}

Textual Data in UPMC are nonverbose and succinct. The challenge in the classification of textual data associated with the UPMC Dataset can be summarized as follows:

\begin{enumerate}
    \item \textbf{Very Short \& Lots of Punctuation: } The data for UPMC Food-101 was collected through web-scraping, and hence, the texts that have been associated with each image are, in actuality, the titles of various images that are available on the web. Therefore, as we generally note from the observation of text data, these titles contain many hyperlinks and punctuation marks. In addition, these texts are shorter than what may actually be required for the model to learn the associations between the text and output labels.

    \item \textbf{Overlapping Texts: } On one hand, the texts are short, and on top of that, these texts in some places (in this dataset) even have overlapping texts. This means that a text may contain the name of another category, whereas it may belong to an entirely different category. This makes it difficult for the model to find appropriate associations between the text and output label.
    
\end{enumerate}
These challenges and issues have been explained in detail with examples from the dataset in Table \ref{analysis-text-models}.

\begin{table*}
\centering
\caption{Identification and study of potential factors that led to degradation in performance of the text classification model. These factors contribute the most to the performance degradation of the textual models.}
\vspace{0.1in}
\label{analysis-text-models}
\resizebox{\textwidth}{!}{
\begin{tabular*}{440pt}{p{90pt}p{90pt}p{200pt}}
\toprule
Text &Actual Class &Issues\\
\midrule
Eat Your Books Mobile & beef\_carpaccio & There is no clear association between the text and actual class. In fact, many samples from 
Some low-performing classes do not have the required keywords that can potentially be used by the model to identify the actual class.\\
\midrule
Steak Tartare & Beef \_ Tartare & There are many examples where the names of other classes are used. In such scenarios, there is no way for the model to correctly associate an example with an appropriate class. Here, the example belongs to beef\_tartare, but the word steak (which is the name of another class) has also been mentioned with no other hint of an expected correct classification. Therefore, as explained earlier, overlapping texts are an issue in the UPMC Food-101 dataset.\\
\midrule
Copa Vida & Croque \_madame & There are many samples in which the text length is too short to be interpreted robustly by text classification models. \\
\bottomrule
\end{tabular*}
}
\end{table*}

Table \ref{Text-distribution-ensemble-original} lists the number of classes with a precision value within a definite range. It can be concluded that the majority of classes have a precision value greater than 0.80, texts belonging to 48 classes have a precision that lies within the range of 0.80 - 0.90, and 38 classes have a precision in the range of 0.90 - 1.00.

\begin{table}
\centering
\caption{Distribution of classes based on the precision values obtained on the original dataset using text classification which employs a voting ensemble}
\vspace{0.1in}
\label{Text-distribution-ensemble-original}
\begin{tabular}{ccc}
\toprule
Range of Precision & Number of classes & Percentage \\
\midrule

0.40 - 0.50 & 1 & 0.99  \\
0.50 - 0.60 & 1 & 0.99 \\
0.60 - 0.70 & 2 & 1.98 \\
0.70 - 0.80 & 11 & 10.89  \\
0.80 - 0.90 & 48 & 47.53 \\
0.90 - 1.00 & 38 & 37.62  \\

\bottomrule
\end{tabular}
\end{table}

From Table \ref{Text-distribution-bert-original}, it can be concluded that texts belonging to approximately 53.46\% of the total classes are classified with a precision that lies within the range 0.80-0.90, and approximately 28.71\% are classified with a precision of 0.90-1.00.

\begin{table}
\centering
\caption{Distribution of classes based on the precision values obtained on the original dataset using text classification which employs BERT}
\vspace{0.1in}
\label{Text-distribution-bert-original}
\begin{tabular}{ccc}
\toprule
Range of Precision & Number of classes & Percentage \\
\midrule

0.40 - 0.50 & 0 & 0.00  \\
0.50 - 0.60 & 2 & 1.98 \\
0.60 - 0.70 & 5 & 4.95 \\
0.70 - 0.80 & 11 & 10.89  \\
0.80 - 0.90 & 54 & 53.46 \\
0.90 - 1.00 & 29 & 28.71  \\

\bottomrule
\end{tabular}
\end{table}

It is evident that some classes performed exceptionally well in terms of precision and recall, whereas others did not. In addition, we are over-optimistic if we look only at the results and do not analyze the underlying causes of the poor performance of our model in some cases. It is important to note here that we compared both machine learning-based and deep learning-based models for text classification. The experimental results indicate that while BERT has higher accuracy on UPMC Food-101, the Voting Ensemble method has higher accuracy on UPMC Food-25. However, we considered a model that performs better for larger datasets, which are likely to be more efficient in handling diverse data. 

The Image part of the dataset is noisy and random. Therefore, we observed a dip in performance, even with the use of highly efficient models. Table \ref{distribution-effMish-original} presents the number of classes with a precision value within a definite range. It can be concluded that after image classification, images belonging to approximately 34.65\% of the classes have been classified with a precision that lies within the range of 0.60-0.70, and approximately 23.76\% are classified in such a manner that the precision lies in the range of 0.70-0.80. We can conclude that there are many variations in the precision and recall for every class. This may be attributed to the varied nature of the classes. We observed that the distribution of images in the training set had a strong influence on the performance during the testing phase. We validated this proposition by observing and analyzing the training images of the worst-, average-, and best-performing classes in terms of precision and recall.

\begin{table}
\centering
\caption{Distribution of classes based on the precision values obtained on the original dataset by the proposed method for image classification}
\vspace{0.1in}
\label{distribution-effMish-original}
\begin{tabular}{ccc}
\toprule
Range of Precision & Number of classes & Percentage \\
\midrule
0.20 - 0.30 & 1  & 0.99 \\
0.30 - 0.40 & 1 & 0.99 \\
0.40 - 0.50 & 13 & 12.87 \\
0.50 - 0.60 & 19 & 18.88 \\
0.60 - 0.70 & 35 & 34.65  \\
0.70 - 0.80 & 24 & 23.76  \\
0.80 - 0.90 & 7 &  6.93 \\
0.90 - 1.00 & 1 &  0.99 \\
\bottomrule
\end{tabular}
\end{table}

A comparison of the visual and textual classifications with state-of-the-art methods is presented in Table \ref{comparison}. From the results, it can be noted that the proposed approach performs significantly better than the other state-of-the-art methods. In terms of accuracy, our method outperformed the second-best performing method by 11.57\% and 6.34\% for image and text classification, respectively.

\begin{table}
\centering
\caption{ Comparison of classification accuracies (Image, Text, and Fusion modalities) across state-of-the-art models on the original UPMC Food-101 dataset. The proposed Dynamic Fusion approach achieves the highest performance across all modalities.}
\vspace{0.1in}
\label{comparison}
\begin{tabular}{cccc}
\toprule
Model Name & Image & Text & Fusion \\
\midrule
ViT \cite{dosovitskiy2020image}   & 69.17 & - & - \\
InceptionV4 + RoBERTa \cite{saklani2024ameliorating}    &   70.59              &   84.90   & - \\
CLIP \cite{radford2021learning}  & - & - & 74.87 \\
FILIP \cite{yao2021filip} & - & - & 75.10 \\
UniCL \cite{yang2022unified}  & - & - & 75.53 \\
iCLIP \cite{wei2023iclip} & - & - & 75.72 \\
FMiFood (without Augmented Text) \cite{pan2024fmifood} & - & - & 76.06 \\
FMiFood (with Augmented Text) \cite{pan2024fmifood} & - & - & 76.22 \\
AlexNet+ Embedded Text (CNN)  \cite{gallo2018image}       & 42.01          & 79.78            & 82.90 \\
Googlenet + Embedded Text(CNN) \cite{gallo2018image}     & 55.65          & 79.78            &  83.37\\
VGG19+TF-IDF  \cite{wang2015recipe}       & 40.21          & 82.06            &  85.10\\
GoogleNet + Word2Vec  \cite{nawaz2018learning}  & 55.65          & 56.75            &  85.69\\
ResNet + FastText \cite{kiela2018efficient}      &   - & - & 90.80 \\
PMF \cite{li2023efficient}      & - & - &  91.51\\
PMF-Large \cite{li2023efficient} & - & - & 91.68\\
HUSE \cite{narayana2019huse}                             & \textbf{73.80}          & 87.30              &  92.30 \\
InceptionV3 + BERT-LSTM \cite{gallo2020image} & 71.67 & 84.41 & 92.50 \\
SVM Linear Stacking Model \cite{suresh2024stacking} & - & - & 93.10 \\
Proposed Work ( Dynamic Fusion)      & 73.60 &  \textbf{88.42}  & \textbf{97.84}  \\
\bottomrule
\end{tabular}
\end{table}

\section{Conclusion}
\label{sec:Conclusion}
Food classification systems are important for various applications that can be used in daily life. Work in this domain is very limited, specifically on the UPMC Food-101 dataset, owing to the requirement of high computational resources to train the network on a large number of classes and data samples. The major drawback of the dataset is the presence of noise and imperfect modalities, owing to its data acquisition process through web scraping.  In this study, we proposed a multimodal classification framework that uses a modified version of EfficientNet with the Mish activation function for image classification, and a traditional BERT transformer-based network is used for text classification. We propose a dynamic multimodal fusion approach that dynamically balances the contributions of each modality, particularly in the presence of noisy or low-quality data. The proposed framework advances computational imaging through a dynamic fusion strategy that adaptively integrates visual and textual features, thereby overcoming challenges such as noisy and incomplete modalities. By significantly enhancing food classification accuracy, our method not only demonstrates robustness in visual data processing but also highlights the broader applicability of adaptive cross-modal fusion for computational imaging tasks. This study paves the way for future research to further optimize imaging techniques and improve multimodal system performance in real-world scenarios. An extensive experimental analysis of the prediction results of both the image and text demonstrated the efficiency and robustness of the proposed approach. The proposed multimodal fusion approach achieved an accuracy of 97.84\% and outperformed existing models. This reiterates that efficient selection and combination of multiple modalities leads to better performance. A detailed analysis can be utilized in the future to further analyze classes with poor performance, and a network can be designed to improve the results of these classes as well. 

\bibliographystyle{elsarticle-num}
\bibliography{References}
\end{document}